%% file: main.tex
\documentclass{INTERSPEECH2023}

\interspeechcameraready

\input{packages}

\input{defenitions}

\title{Approximate Nearest Neighbour  Phrase Mining \\ for Contextual Speech Recognition}

\name{
Maurits Bleeker$^{1^*}$\thanks{$^*$Work done while interning at Apple.}, Pawel Swietojanski$^2$, Stefan Braun$^2$, and Xiaodan Zhuang$^2$}
\address{
  $^1$University of Amsterdam, The Netherlands\\
  $^2$Apple}
\email{m.j.r.bleeker@uva.nl, \{pswietojanski, stefan\_braun, xiaodan\_zhuang\}@apple.com}

\begin{document}

\maketitle
 
\input{sections/00_abstract}

\input{meta}

\input{sections/01_introduction}
\input{sections/02_method}
\input{sections/03_experimental_setup}
\input{sections/04_results}
\input{sections/05_conclusion}

{
\small 
\section{Acknowledgements}
We thank Dogan Can, Nikos Flemotomos, Roger Hsiao, Vijay Peddinti, Honza Silovsky, and Russ Webb for their valuable suggestions on this work.
}

\bibliographystyle{IEEEtran}
\bibliography{references}

\end{document}

%% file: packages.tex
\usepackage{amssymb} 
\usepackage{algorithmic}
\usepackage{graphicx}
\usepackage{textcomp}
\usepackage{xcolor}
\usepackage{hyperref}       
\usepackage{url}            
\usepackage{booktabs}       
\usepackage{amsfonts}       
\usepackage{nicefrac}       
\usepackage{microtype}      
\usepackage{multirow}
\usepackage{subcaption}
\usepackage{newfloat}
\usepackage{floatrow}
\usepackage{float}
 \usepackage{soul}
\usepackage{acronym}
\usepackage[sort,nocompress]{cite}
\usepackage{pifont}
\usepackage{caption}

%% file: defenitions.tex
\newcommand{\xmark}{\ding{55}}%

\newacro{ANN}{approximate nearest neighbour}
\newacro{ANN-P}{approximate nearest neighbour phrases}
\newacro{ANN-PI}{approximate nearest neighbour phrases index}
\newacro{HNP}{hard negative phrases}
\newacro{E2E}{end-to-end}
\newacro{ASR}{automatic speech recognition}
\newacro{CT}{Conformer Transducer}
\newacro{CCT}{Contextual Conformer Transducer}
\newacro{MHA}{multi-head attention}
\newacro{CATT}{Context-Aware Transformer Transducer}
\newacro{CAT}{Context-Aware Transducer}
\newacro{CLAS}{Contextual Listen, Attend, and Spell}
\newacro{RNN-T}{RNN Transducer}
\newacro{LAS}{Listen-Attend-Spel}
\newacro{WER}{Word Error Rate}
\newacro{LM}{language model}
\newacro{DCF}{deep neural contextual fusion}

%% file: sections/00_abstract.tex
\begin{abstract}
This paper presents an extension to train \acl{E2E} \acf{CATT} models by using a simple, yet efficient method of mining hard negative phrases from the latent space of the context encoder.
During training, given a reference query, we mine a number of similar phrases using approximate nearest neighbour search.
These sampled phrases are then used as negative examples in the context list alongside random and ground truth contextual information. 
By including \acl{ANN-P} in the context list during training, we encourage the learned representation to disambiguate between similar, but not identical, biasing phrases. This improves biasing accuracy when there are several similar phrases in the biasing inventory.
We carry out experiments in a large-scale data regime obtaining up to 7\% relative word error rate reductions for the contextual portion of test data.
We also extend and evaluate \ac{CATT} approach in streaming applications. 
\end{abstract}

%% file: meta.tex
\noindent\textbf{Index Terms}: Neural Transducer, Contextual Speech Recognition, Approximate Nearest Neighbour sampling

%% file: sections/01_introduction.tex
\section{Introduction}
\label{sec:intro}

Recognizing words that are rare or unseen during training poses a challenge for \acf{E2E} \acf{ASR}~\cite{guo2019spelling, sainath2018no, bruguier2019phoebe, alon2019contextual}. 
One way to address this problem is to allow the model to use user-specific information during inference, such as contact names, app names, media titles, and relevant geo-location names. 
To that end, several approaches have been proposed including shallow language model (LM) fusion~\cite{zhao2019shallow, le2021deep}, on-the-fly rescoring~\cite{hall2015composition, williams2018contextual, zhao2019shallow}, or deep fusion approaches~\cite{pundak2018deep, bruguier2019phoebe, jain2020contextual}. 
As \ac{E2E} models tend to learn a strong internal LM~\cite{mcdermott2021dr, meng2021internal}, shallow LM fusion and rescoring approaches are not always effective out of the box.

Alternative methods rely on deep neural contextual fusion (DCF)~\cite{pundak2018deep, chang2021context, sathyendra2022contextual, bruguier2019phoebe, jain2020contextual, sun2021tree, munkhdalai2022fast}. 
In DCF, the biasing machinery is part of the \ac{ASR} model and is jointly learned with the main \ac{ASR} objective. 
Different DCF techniques share much of the same modeling back-end and thus can be implemented for arbitrary \ac{E2E} network architectures such as the attention encoder-decoder (AED)~\cite{chan2015listen} or the neural transducer~\cite{graves2012sequence} (RNN-T)~\ac{ASR} systems. Deep contextual biasing has been proposed for the \ac{CLAS}~\cite{pundak2018deep, chan2015listen} AED model, and similar solutions were extended to the contextual neural transducers~\cite{jain2020contextual, chang2021context}. 
The major difference between contextual models and their non-contextual counterparts is the biasing machinery, usually implemented as an additional context encoder followed by a fusion mechanism. 
The context encoder is typically implemented as an LSTM~\cite{hochreiter1997long} or more recently a transformer~\cite{vaswani2017attention} model, and its role is to project a set of tokenized biasing phrases into a set of fixed-sized continuous embeddings. 
Next, a fusion mechanism integrates these embeddings with the acoustic (AED, RNN-T) and/or label (RNN-T) encoder when making \ac{ASR} predictions. 
Fusion can be implemented in a latent space with cross-attention between audio and context encoders~\cite{pundak2018deep, jain2020contextual} or by interpolating generic and contextual model's distributions, as done in tree-constrained pointer generation networks (TCPGN)~\cite{sun2021tree}.

A major challenge in contextual biasing is that some words, including the biasing phrases fed into the context encoder, may exhibit phonetic similarities with one another or may be characterized by complex and non-standard pronunciation patterns.
For example, names in a contact list that sound similar to each other, or  geo-location names that have similar (but not identical) pronunciations. To make deep contextual biasing more robust to settings where context information is (phonetically) similar to each other, one could explicitly embed additional phoneme-level information in the contextual ASR as explored in~\cite{chen2019joint}, or train the \ac{ASR} system such that it can learn to better disambiguate between challenging queries.

In this work, we are interested in the latter approach by exposing the ASR to hard negative examples during training.
Alon et al.~\cite{alon2019contextual} proposed a method to generate phonetically similar phrases given a reference phrase.
Phonetically similar phrases might have similar (i.e.,~confusing) acoustic representations and, therefore, are hard to distinguish from the desired phrase during inference~\cite{alon2019contextual}.
By appending phonetically similar phrases as~\textit{hard negatives} to the context encoder's inputs during training, the model is explicitly tasked to disambiguate between them. 
There are several possible ways to insert hard negatives into the training pipeline.
In ~\cite{alon2019contextual}, an external \ac{ASR} model~\cite{variani2017end} is used to decode and generate a set of hypotheses for each query. 
These hypotheses are then ranked based on the word co-occurrence and the phonetic similarity with the reference phrase. 

While the method by Alon et al.~\cite{alon2019contextual} has shown promising results, it is worth noting that their approach may be viewed as a form of data augmentation implemented prior to training of a deep contextual \ac{ASR} model, rather than a technique that can be directly integrated into the training process. 
This may not be optimal, as exact \ac{HNP} are likely to depend on the mistakes a specific \ac{ASR} is prone to make, rather than mis-recognitions of some independent ASR system. 

In this paper, we present an alternative, computationally efficient extension of mining \ac{HNP}: \acf{ANN-P} mining. 
\ac{ANN-P} mining allows to efficiently select \ac{HNP} during the training of a deep contextual model in an online manner, using the latent space of the context encoder.
\ac{ANN-P} mining unifies two important aspects of \ac{HNP} for contextual \ac{ASR} in a single method, by including phrases in the context list that are
\begin{enumerate*}[label=(\roman*)]
\item phonetically similar to the reference phrase (i.e., distracting phrases),
as in \cite{alon2019contextual}, and
\item close to the reference phrase in the latent space of the context encoder. The latter property is hypothesized to make it harder for the model to discriminate different phrases \cite{alon2019contextual}.
\end{enumerate*}
Different from Alon et al.~\cite{alon2019contextual}, our approach does not require a full decode of training data, nor an existing pre-trained \ac{ASR} model to obtain hard negatives.

We implement the proposed \ac{ANN-P} mining using the \acf{CATT} model~\cite{chang2021context}. 
\ac{CATT} proposed an efficient biasing approach that makes full use of the Transformer Transducer (TT)~\cite{Zhang2020} architecture. 
The biasing phrases in \ac{CATT} are mapped into single embeddings (one per biasing phrase) that is then used with cross-attention to bias both audio and label encoders. 
While the original \ac{CATT} formulation only considered non-streaming scenarios, we extend CATT to both streaming and non-streaming applications at the same time by training it in a variable attention masking manner~\cite{Anshuman2020, var_masking}.
 Given the limited (audio) context window in streaming scenarios, it is difficult to accurately distinguish the correct biasing phrase while transcribing. This especially applies for CATT-like approaches where each biasing phrase is compressed into a single embedding, rather than an embedding per sub-word as is the case with TCPGN or neural associative networks (NAM)~\cite{munkhdalai2022fast}
 \footnote{Although not investigated in this work, CATT is likely to work well for neural biasing of non-auto-regressive models like connectionist temporal classification (CTC)~\cite{Graves2006icml}, whereas NAM or TCPGN relies on decoded prefixes for biasing. See~\cite{dingliwal2023personalization} for a recent CTC study.}.
 As such, we anticipate that for the \ac{CATT} model, \ac{ANN-P} mining is likely to provide greater value in streaming than in non-streaming scenarios.

The contributions of this work thus include:
\begin{itemize}
 \item An extension of the training mechanism for CATT that utilizes mining hard negative phrases directly from the latent space of the \ac{ASR} model. This approach results in up to 7\% relative word error rate reductions for the personalized portion of test data, with only minor regressions on generic queries.
 \item An extension of CATT to the streaming scenario.
 \item Evaluation of the proposed approach in a large-scale data regime consisting of 650,000 hours of acoustic training data. This is to make sure the models are well-trained, and not handicapped by limited training data diversity, which in turn could artificially inflate biasing performance.
\end{itemize}

%% file: sections/02_method.tex
\begin{figure}[!tb]
    \centering
    \includegraphics[width=1.0\textwidth]{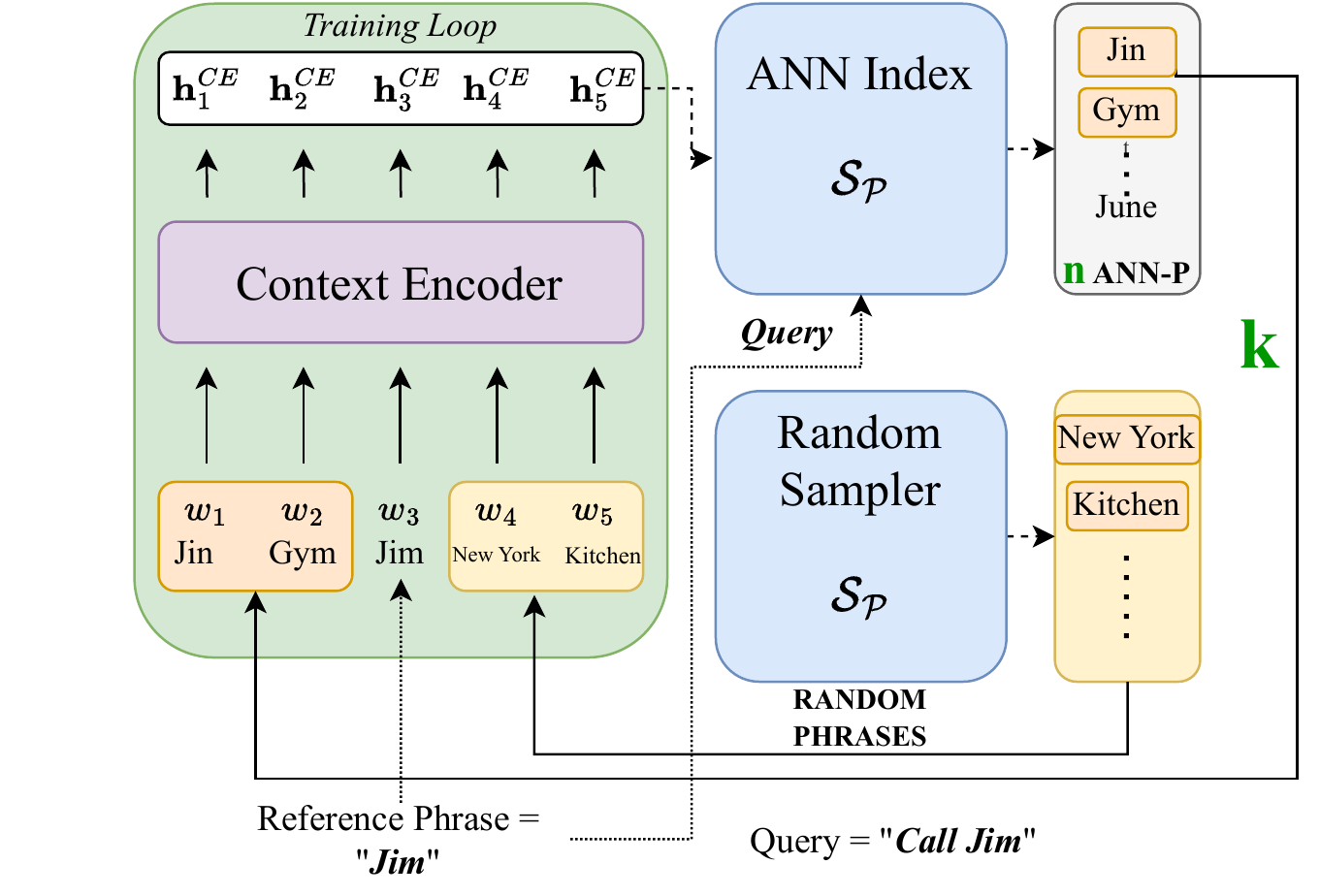}    
    \caption{Overview of the \ac{ANN-P} mining method.  Given a query \textbf{``Call Jim''}, we append \textit{k} ANN phrases to the context list. Given a query phrase, we first sample $n$ phrases from the ANN index. Next, we sample the top $k$. The remaining phrases in the context list are randomly sampled.
    \vspace{-0.1cm}
    }
    \label{fig:method}
\end{figure}

\section{Method}
\label{sec:method}

\subsection{Context-Aware Transformer Transducer}

\ac{CATT}~\cite{chang2021context} extends an RNN-T consisting of a label and audio encoder, with an additional contextual encoder, followed by a biasing cross-attention layer that measures the relevance of the context phrase to the query from the perspective of information found in audio and label encoders.
The context encoder~ $f^{context}(\cdot)$ takes as input a set of context phrases $\mathcal{S}_{\mathcal{C}} = \{w_{1}, \dots\}$ and maps each context phrase $w_{i} \in \mathcal{S}_{\mathcal{C}}$ into a fixed vector representation $f^{context}(w_{i}) := \mathbf{h}_{i}^{CE}$.
CATT uses a transformer~\cite{vaswani2017attention} for the label, audio, and context encoder. 

The biasing layer in the \ac{CATT} model consists of a \ac{MHA} layer~\cite{vaswani2017attention}. 
The goal of the \ac{MHA} is to measure the similarity between each phrase in the context list and the audio signal using scaled dot-product cross-attention, which weights each input phrase according to this similarity score with the audio.

Consider a pair of contextual phrases $w_i, w_j \in \mathcal{S}_{\mathcal{C}}$, where $i \neq j$.  
Phrase $w_{i}$ is the reference phrase in the audio signal and $w_{j}$ is considered as a negative phrase w.r.t. the audio transcript.
Phrases $i$ and $j$ have close to each other (key) embeddings ($\mathbf{k}_{i}$ and $\mathbf{k}_{j}$ respectively), in terms of a similarity metric. 
When a query-key pair has a low attention score, the matching value embeddings get close to zero weight score.
Since $\mathbf{k}_{i}$ and $\mathbf{k}_{j}$ are close to each other, $\mathbf{k}_{j}$ may be distracting for the attention head(s) yielding too high attention score, hence the distracting phrase may be taken into account when transcribing the audio.

\subsection{Proposed method: ANN-P mining}

In this section, we present an approach to extend the training of the \ac{CATT} by introducing \ac{ANN-P} mining.
The goal of \ac{ANN-P} mining is to select phrases similar to the \textit{reference phrase} in terms of their similarity in the latent space of the context encoder.
When mining the phrases randomly, the probability of having (phonetically) similar phrases in the context list is negligible.
Therefore, the label, audio, and context encoder may not learn to disambiguate between similar sounding (i.e., difficult to discriminate) phrases. 
In Fig.~\ref{fig:method} we provide a high-level overview of our \ac{ANN-P} mining method.

Prior to training, we extract all the biasing phrases from each audio transcription using the existing automatically generated meta-information on entity spans. 
This results in a set of phrases for the entire training data $\mathcal{S}_{\mathcal{P}}$, referred to as the biasing phrases inventory.
$\mathcal{S}_{\mathcal{P}}$ can be extended with additional entries that are not present in the training data to provide additional context, as needed. 

The goal of \ac{ANN-P} mining is to select hard negative samples according to the current state of the trained \ac{ASR} model.
We use the context encoder of the \ac{CATT} to encode each phrase $w_{i} \in \mathcal{S}_{\mathcal{P}}$ into its latent representation $\mathbf{h}^{CE}_{i}$ given a checkpoint of the \ac{CATT} during training (left box in Fig.~\ref{fig:method}).
We cache the latent representation $\mathbf{h}_{i}^{CE}$ of each phrase into an online \ac{ANN} index (i.e., an index that can be efficiently queried during training). 

Given a \textit{query phrase} $w_{i}$, the \ac{ANN} index maps the phrase $w_{i}$ to its cached latent representation $w_{i} \rightarrow \mathbf{h}^{CE}_{i}$ and, using \ac{ANN} search over all cached phrase embeddings, returns $n$ \ac{ANN-P} from the index based on the dot product score w.r.t. the query phrase.
To prevent sampling the same phrases at every epoch for the same query, we randomly sample $k$ (where $k < n$) phrases from the $n$ retrieved phrases.
The remaining phrases (if any) that are needed for the given query are added randomly by sampling from $\mathcal{S}_{\mathcal{P}}$.
Together the \ac{ANN-P} and the randomly sampled phrases are included in the context list.
\ac{ANN-P} mining can only be used if the query phrase is present in the \ac{ANN} index since we need its neural representation to apply similarity search.
Hence, we need to cache each phrase first before we can apply \ac{ANN-P} mining. 
However, an indexing step is an inexpensive process, taking a small percentage of the total training time.

For \ac{ANN-P} mining, we need one representation per biasing phrase (all from the same latent space) to store in the ANN index. 
In this work, we use the output representation of the context encoder $\mathbf{h}^{CE}_{i}$ for \ac{ANN-P} mining. 
However, \ac{CATT} measures the similarity between the audio and the phrases by using a {MHA} biasing layer, which uses 8 attention heads. 
Hence, there are 8 different phrase representations (living in different latent spaces) that are used to compute an attention score (i,e, similarity) between a phrase and the audio signal. 
There are several ways to aggregate the 8 key representations into a single representation that could be used for approximate nearest neighbour search.  
As a straightforward approach that applies as an approximation for all the 8 key representations, we store the output representation of the context encoder $\mathbf{h}^{CE}_{i}$ in the index instead (this is because the key projection is only a linear transformation).

%% file: sections/03_experimental_setup.tex
\label{sec:experimental}
\section{Experimental Setup}

We carry out the experiments on a large-scale dataset consisting of queries from two tasks: dictation and assistant. 
The semi-supervised portion of the data consists of around 600,000 hours of randomized and anonymized automatically transcribed acoustic data, while the supervised part contains around 50,000 hours of randomized and anonymized English queries. 

Following \cite{var_masking, nguyen2022optimizing}, our systems are trained in a two-stage manner - the first stage pre-trains the models on semi-supervised data for a total of 5.6M updates, the second fine-tunes the model for another 280k updates on supervised data. 
In both stages, gradients are accumulated over 9216 queries. 
We use SyncSGD + Adam~\cite{kingma2014adam} for distributed optimization, with exponentially decaying learning rates. 
\ac{ANN-P} sampling is applied only in the fine-tuning stage and not during evaluation. 
All models are evaluated using a test set containing 60 hours of assistant data. 
Around 40\% of the test set consists of contextual queries spanning domains such as contact, app, and geo-location names. 
During inference we include real user profiles in the context list.
The remainder consists of queries that are generic in nature and are unlikely to benefit from personalized priors.

\subsection{Contextual Transformer Transducer Model}

In this work, our base contextual \ac{E2E} \ac{ASR} model is the \acf{CATT}~\cite{chang2021context}, configured to have around 120M parameters. 
The audio encoder is a 12-layer Conformer~\cite{gulati2020conformer} while the label and context encoders are implemented as a 6-layer transformer model. 
Each encoder has an embedding size of 512 and the \ac{MHA} is configured to 8 attention heads.
All examples in the training batch share the same context list, which allows exposing each query to a larger number of context phrases, at the same time keeping the memory usage low. 
To do so, we only sample a few random + \ac{ANN-P} per query and combine them into a single context list for all queries in a mini-batch. 
This was configured such that each query has access to around 96 - 128 biasing phrases.

Different from the original \ac{CATT}, we append a back-off phrase to the context list where the model can attend to in case there are no relevant-biasing phrases. 
Adding a back-off token has also been demonstrated to be effective with CLAS~\cite{pundak2018deep}.
Originally, CATT was only trained and evaluated with global context models.
In this work, we also investigate its suitability to both non-streaming and streaming applications. 
We do so by training both CATT and baseline models in variable masking manner\cite{Anshuman2020, var_masking}, and then configuring the models to either streaming or non-streaming settings during decoding.
Streaming models operate on 240ms long causal audio chunks~\cite{shi2021emformer, Chen2021} and thus have limited access to the future audio signal which may pose a challenge for CATT approach.

To show performances without biasing machinery, we compare with a Transformer Transducer (TT)~\cite{Zhang2020} that has the same audio and label encoder architecture as CATT and has been trained with multiple attention masks to allow for streaming and non-streaming decoding. 
The exact architecture details of the TT can be found in~\cite{var_masking}. 
Since streaming models are expected to emit tokens with low partial latency, we train both TT and CATT models with the latency-penalizing FastEmit loss~\cite{yu2021fastemit}. 

\subsection{ANN index and negative phrase mining}

The \ac{ANN-P} index is built using the Annoy\footnote{\url{https://github.com/spotify/annoy}} library. 
For the \ac{ANN-P} mining, we experiment with various ways of mixing negative examples into context lists. 
In general, we append 8 biasing phrases for each query\footnote{Note, we eventually share these across all examples in the batch, so for a batch of 16 queries and 8 contextual phrases per query, each query would make use of 128 biasing phrases to pick from.}
, where some proportion is expected to be made of \ac{ANN-P} (see section \ref{sec:further} for details), while the remainder is randomly selected out of the biasing phrases inventory. 
We use the dot product between phrase embeddings and a query as 
similarity metric to mine \ac{ANN-P}.
Given a training query, and its corresponding reference biasing phrase(s) (if any), we retrieve \emph{n} approximate nearest neighbours form the index.  
Important to notice, we sample each negative phrase at the word level (i.e., if a query phrase consists of multiple words, we sample $n$ \ac{HNP}s per word). 
Next, we sample \emph{k} phrases at random from the retrieved \emph{n} \text{ANN}s. 
After completing two fine-tuning epochs, the \ac{ANN} index is rebuilt by re-indexing the phrase representations using the latest state of context encoder parameters. 
To prevent over-fitting on the same \ac{ANN-P}, we do not sample \ac{ANN-P} for every query in the training batch but use them proportionally to the \textit{append ratio}. 
The frequency of adding \ac{ANN-P} to the context list during training increases as the append ratio increases.
In cases where we are unable to sample \ac{ANN-P}s for a query, we use random phrases instead. 

%% file: sections/04_results.tex
\input{tables/result_table.tex}

\section{Results}
\label{sec:results}

\subsection{Random vs ANN-sampled phrases}

Table~\ref{table:results} reports the results for models operating in global (i.e., non-streaming) (upper block) and streaming (lower block) modes, with and without access to contextual information. 
To demonstrate the effect on \ac{WER} in situations where contextual information is absent, we also evaluate \ac{CATT} without access to relevant contextual information.
In this work, global and streaming models are the same models, configured to different operating regimes via different settings of attention masking~\cite{var_masking}. 
Note that~\cite{chang2021context} only investigated the biasing performance of CATT in a global setting, and thus it is unclear if and to what extent the CATT approach can be used in streaming applications. 

When decoding CATT models in global mode, allowing the model to access contextual information during inference improves accuracy by 38\% relative WER (WERR) on average (i.e., 6.5\% vs. 3.9\% for non-biased and biased CATT systems, respectively).
This is accompanied by a 3\% WERR degradation on a non-personal (generic) portion of the test set (i.e., 6.3\% vs. 6.5\% on average for TT and CATT, respectively). 
These results are in line with findings on CATT and global decodes reported in~\cite{chang2021context}.
Another observation is that training with \ac{ANN-P} mining does not seem to affect the non-streaming results in a significant way. 
This can be most likely explained by the fact that having access to the entire audio sequence allows the model to better contextualize the information, thus it is easier to match the complete audio evidence to the correct biasing phrase.

When decoding CATT models configured to streaming mode, we observe similar overall trends as with the global models. 
Interestingly, for the streaming scenario, the regression on WER for the non-contextual portion of data no longer exists when compared to a baseline TT model (i.e., 6.8\% vs 6.7\% for the baseline and CATT, respectively). 
For the streaming scenario, we also obtain up to 7\% WERR reductions on the contextual portion of the test set (this is where \ac{ANN-P} is expected to help). 

We can conclude that for the limited audio look-ahead, training \ac{CATT} with \ac{ANN-P} mining helps to improve accuracy by allowing the model to better disambiguate between contextual phrases. 
Since in CATT the biasing information is compressed into a single embedding, the use of \ac{HNP} helps to regularize embeddings such they are more robust to small phonetic variations.

\subsection {Further analyses}\label{sec:further}

\begin{figure}[h!]
	\centering
	\includegraphics[width=0.9\linewidth]{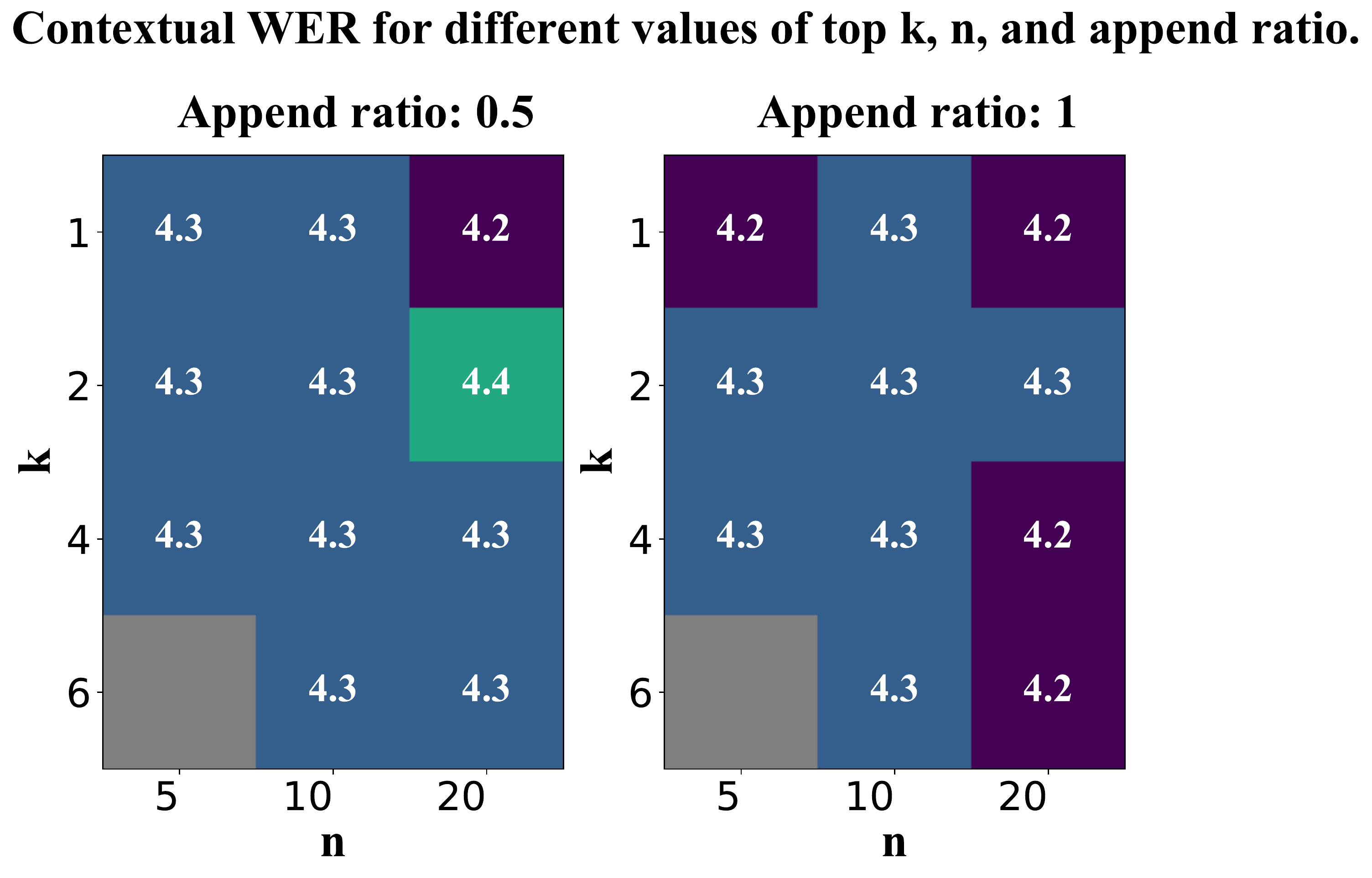}
    \vspace{-0.5cm}
	\caption{Average WER for streaming decodes using different values of top k, n, and append ratio.
    \vspace{-0.35cm}
    }
	\label{fig:hyper_paras}
\end{figure}

\ac{ANN-P} mining has three main hyper-parameters: \textit{n} (i.e, the number of \ac{ANN-P} we take from the index), \textit{k} (i.e., the $k$ phrases we sample from the \textit{n} ANNs), and the \textit{append ratio} (how frequent we add the \ac{ANN-P} to the context list). We depict the effect of each parameter in Fig.~\ref{fig:hyper_paras}. 
We can conclude that, in general, \ac{ANN-P} mining is robust to the choice of the considered hyper-parameters.
The lowest scores are obtained for an append-ratio of 1 and using $n=20$ phrases from the index. 
The number of phrases ($k$) appended to the context list does not seem to have a strong effect on the \ac{WER}.

\input{tables/query_phrases.tex}

In this work, we mine hard negative phrases (at the word level) from the latent space of the context encoder, based on the neural similarity with the reference (i.e., query) phrase. 
In Table~\ref{table:query_ann}, we provide several examples of query phrases and their top four \ac{ANN-P} as retrieved from the index. 
We can observe that top \ac{ANN-P} mainly results in phrases that are phonetically similar to the query phrase.
For queries consisting of names, we mainly retrieve other similar-sounding (but different) names.
For a query such as \textit{building}, we mainly retrieve \ac{ANN-P} that are related to the same concept and contain the sub-word \textit{build}.

Finally, we also investigated the following aspects and report them here for completeness. 
These experiments either did not significantly impact accuracy or led to deterioration:
\begin{itemize}
\item Rebuilding the \ac{ANN-P} index at different epochs did not have a significant impact on WER. 
\item Enabling \ac{ANN-P} mining at different stages of training, including the pre-training, or fine-tuning the last 2-3 epochs did not improve over using it during the entire fine-tuning stage.
\item Sampling \ac{ANN-P} using multi-word phrases, instead of single-word phrases, resulted in 5\% WERR degradation.
\end{itemize}

%% file: tables/result_table.tex
\begin{table}[!tb]

\begin{tabular}{c|c||c|c|c}
\multirow{2}{*}{Model}  & \multirow{2}{0.8cm}{\centering Ctx. Info} & \multicolumn{3}{c}{WER [\%]} \\ \cline{3-5}
                        &                  & Generic & Personal & Avg. \\ \hline \hline
\multicolumn{5}{c}{Global decoding}     \\ \hline       
TT                 & -                & 3.9 & 11.9 & 6.3 \\ \hline
CATT                    & \xmark                & 4.3 & 11.5 & 6.5 \\ 
CATT w/ ANN-P           & \xmark                & 4.4 & 11.5 & 6.5 \\ \hline
CATT                    & \checkmark       & 4.5 & 2.6 & 3.9 \\ 
CATT w/ ANN-P           & \checkmark       & 4.5 & 2.6 & 3.9 \\ \hline \hline
\multicolumn{5}{c}{Streaming decoding}     \\ \hline
TT                 & -                & 4.5 & 12.3 & 6.8 \\ \hline
CATT                    & \xmark                & 4.6 & 11.4 & 6.7 \\ 
CATT w/ ANN-P           & \xmark                 & 4.5 & 11.6 & 6.7 \\ \hline
CATT                    & \checkmark       & 4.9 & 2.8 & 4.3 \\ 
CATT w/ ANN-P          & \checkmark       & 4.9 & \bf{2.6} & \bf{4.2} \\ \hline \hline
\end{tabular}
\caption{WERs for models configured to global (upper block) and streaming (lower block) decodings, with and without access to contextual information (Ctx. Info). CATT models are trained with random, or \ac{ANN-P} sampling. The results on test data are additionally aggregated on generic and personalized subsets. The Generic portion consists of queries that are not expected to benefit from contextual information. Results obtained for $n$=20, $k$=2, and append ratio=0.25.
}
\label{table:results}
\end{table}

%% file: tables/query_phrases.tex
\begin{table}[!tb]
\begin{tabular}{c|l}
Query Phrase & \multicolumn{1}{c}{$n=4$ ANN Phrases}                                                                                                                         \\ \hline
john     &  'joan', 'johnson', 'johann', 'from john'                                                                                                                      \\
building     & 'buildings', 'builder', 'the building', 'builds'                                                                                                              \\
jean         & 'jeanne', 'jeannie', 'jeana', 'jeanine'                                                                                                                       \\
eva  & \begin{tabular}[c]{@{}l@{}}'evie', 'ava', 'evin', 'evy'\end{tabular}                                              \\
play & \begin{tabular}[c]{@{}l@{}}'playa', 'place', 'flay', 'platte'\end{tabular} \\ \hline
\end{tabular}
\caption{Query Phrases and their ANNs retrieved from the index.
}
\label{table:query_ann}
\end{table}

%% file: sections/05_conclusion.tex
\section{Discussion \& Conclusion}
\label{sec:conclusion}

We proposed and evaluated an efficient method for mining approximate nearest neighbour phrases (i.e., hard negatives) for transformer-transducer contextual speech recognition based on \ac{CATT} model. 
In order to mine hard negatives, our method does not require an external ASR model, nor additional decodings of training data.
We also extended CATT modelling to streaming applications by training it with multiple attention mask configurations.
We evaluated the proposed ideas in large-scale data experiments, finding that the CATT using the ANN-P mining approach offers up to 7\% relative \ac{WER} reductions for streaming models on the personalized portion of the test data.